\documentclass[conference]{IEEEtran}
\ifCLASSINFOpdf
\else
\fi
\usepackage{graphicx}
\usepackage{caption}
\usepackage{balance}

\begin{document}
%
\title{A Brief Review for Compression and Transfer Learning Techniques in DeepFake Detection}

\author{
\IEEEauthorblockN{Andreas Karathanasis, John Violos, Ioannis Kompatsiaris, Symeon Papadopoulos}
\IEEEauthorblockA{Information Technologies Institute, Centre for Research \& Technology Hellas, Thessaloniki, 57001, Greece \\
Email: {andrew.karathanasis@iti.gr, violos@iti.com, ikom@iti.gr, papadop@iti.gr }}}


%


\maketitle

\begin{abstract}
Training and deploying deepfake detection models on edge devices offers the advantage of maintaining data privacy and confidentiality by processing it close to its source. However, this approach is constrained by the limited computational and memory resources available at the edge. To address this challenge, we explore compression techniques to reduce computational demands and inference time, alongside transfer learning methods to minimize training overhead. Using the Synthbuster, RAISE, and ForenSynths datasets, we evaluate the effectiveness of pruning, knowledge distillation (KD), quantization, fine-tuning, and adapter-based techniques. 
Our experimental results demonstrate that both compression and transfer learning can be effectively achieved, even with a high compression level of 90\%, remaining at the same performance level when the training and validation data originate from the same DeepFake model. However, when the testing dataset is generated by DeepFake models not present in the training set, a domain generalization issue becomes evident.
\end{abstract}

\IEEEpeerreviewmaketitle

\section{Introduction}
Deepfake detection often involves computationally intensive deep learning models that process high-dimensional images. Without compression, these models can be prohibitively large and slow, making real-time detection impractical, especially in edge computing environments \cite{violos_towards_2024}. By reducing model size while preserving accuracy, compression techniques allow for efficient inference, lower latency, and reduced energy consumption \cite{kinnas_reducing_2025}, making advanced deepfake detection accessible to a broader range of devices and applications.

Transfer learning can play a pivotal role in deepfake detection by leveraging pre-trained models developed for related tasks, such as image recognition or facial analysis, to save time and computational resources \cite{pan_survey_2010}. Deepfake datasets are often diverse and complex, requiring significant effort to train a model from scratch. Transfer learning allows models to reuse learned features from large, general purpose datasets and adapt them to the specific task of deepfake detection, even with limited labeled data \cite{ayana_multistage_2024}. This approach enhances detection accuracy, accelerates development, and helps tackle the challenge of detecting evolving deepfake techniques.

Although numerous studies explore deepfake detection, model compression techniques, and transfer learning independently, there is a lack of research on combining these approaches in deepfake detection models. To address this gap, we investigate the use of techniques such as pruning, KD, quantization, fine-tuning, and adapters in deepfake detection. Our work includes an analysis of their applications and an experimental evaluation across three datasets.

The structure of this paper is organized as follows: Section \ref{sec:RelatedWork} offers a concise overview of the techniques employed. Section \ref{sec:ProposedMethodology}  details the methodologies explored for achieving compression and transfer learning in DeepFake detection. Section \ref{sec:ExperimentalEvaluation} presents the experimental results along with a discussion of their implications. Finally, Section \ref{sec:Conclusions} outlines our conclusions and directions for future research.

\section{Related Work \& Background}
\label{sec:RelatedWork}
Deepfake detection refers to the identification and classification of manipulated digital media, such as images and videos, that are generated using deep learning techniques like Generative Adversarial Networks and Diffusion Models \cite{rana_deepfake_2022}. These manipulations often include face swaps, facial reenactments, or synthetic generation of human likeness, posing threats to privacy, authenticity, and public trust \cite{wang2020cnn}. The survey \cite{nguyen2022deep} provides a comprehensive analysis of both the creation and detection of deepfakes, emphasizing the challenges posed by their increasing sophistication. Furthermore it explores potential legal implications that could facilitate the reduction of DeepFake material in public media platforms. While deep learning-based approaches can tackle the challenge of DeepFake detection, they typically demand substantial computational resources for both training and testing, making them unsuitable for edge computing applications. At the same time, as highlighted by \cite{mirsky2021creation}, although the creation of deepfakes has become increasingly accessible due to open-source tools and pre-trained models, producing high-quality results still relies heavily on large datasets and, more importantly, significant computational resources. To prevent a potential "war" for resources, the development of resource-efficient detection techniques is imperative.

Techniques such as pruning, KD, and quantization have been proposed for model compression across various tasks and modalities.
Pruning involves reducing the complexity of a neural network by removing less significant parameters, such as weights or neurons, while retaining its predictive power \cite{liang_pruning_2021}. 
KD transfers knowledge from a larger, more complex model (the teacher) to a smaller model (the student) \cite{violos_towards_2024}. The student model learns to mimic the teacher by approximating its predictions or output probabilities \cite{tsanakas_light-weight_2024}. 
Quantization simplifies models by reducing the precision of their numerical representations, such as converting 32-bit floating-point values to 8-bit integers \cite{liang_pruning_2021}. This significantly reduces memory and computational requirements with minimal accuracy loss. 

Fine-tuning is a key technique that can be applied after compressing deep learning models to restore any lost performance and optimize their accuracy \cite{guo_spottune_2019}. Additionally, adapters, which are lightweight, trainable modules inserted into pre-trained models, offer an efficient method for transfer learning by enabling the model to adapt to new tasks or datasets without extensive re-training \cite{sung_vl-adapter_2022}. 
However, determining the most effective approach between these techniques for deepfake detection remains an open question. This research addresses the gap both theoretically in Section 3 and experimentally in Section 4, providing an experimental comparison.

\section{Approaches for Compression \& Transfer Learning}
\label{sec:ProposedMethodology}

We explore a baseline method, three compression and six transfer learning approaches, which are briefly presented below. For a comprehensive overview of these methods, we direct readers to  \cite{karathanasis_comparative_2025}.
The baseline and the compression are applied in one dataset. Transfer learning utilizes two datasets: a) the dataset used to train a model on a general or related task, which is called the \textit{source} dataset, and b) the one used to fine-tune the model for a specific task, i.e. the \textit{target} dataset.

\begin{enumerate}
   \item[(1)] \textbf{Baseline (BL):} The baseline approach involves straightforward training of the original, large deepfake detection model.
    
    \item[(2)] \textbf{Compression with Pruning and Fine-Tuning (CPF):} The first compression approach involves pruning the originally trained BL model , followed by a fine-tuning process. In this approach, the pruned model undergoes training across all layers for a few epochs to reduce the performance degradation caused by pruning.

    \item[(3)] \textbf{Compression with Knowledge Distillation without Fine-Tuning (CKD):} In this approach, the BL model serves as the teacher model, while a smaller model is used as the student model. Notably, no fine-tuning is performed in this process.

    \item[(4)] \textbf{Compression with Quantization (CQ):} In this approach, we quantized the weights of the linear layers of the BL model, reducing their precision from float32 (the PyTorch default) to int8, the smallest supported precision.

    \item[(5)] \textbf{Transfer Learning with Fine-Tuning (TL+FT):} In this approach, the BL model which is trained on the source dataset is fine-tuned on selected layers using the target dataset

    \item[(6)] \textbf{Transfer Learning with Fine Tuning and Quantization (TL+FT+Q):} The transferred large model, undergoes the CQ method and it's subsequently quantized in its linear layers to achieve compression. Fine-tuning facilitates adaptation to the target data, while quantization addresses model compression.
    
    \item[(7)] \textbf{Transfer Learning with Pruning and Fine-Tuning (TL+P+FT):} The pruned models, initially trained and fine-tuned on the source data as outlined in CPF method, are afterwards, further fine-tuned on the target data, in some selected convolutional layers. The pruning process addresses model compression, while the additional fine-tuning on the target data enables transfer learning.

    \item[(8)] \textbf{Transfer Learning with Knowledge Distillation (TL+KD):} In this approach, the BL model serves as the teacher model, aiming to transfer general, low-complexity knowledge learned from the source data that is also relevant to the target data. The student model is a smaller, completely untrained model, and the target data is used during the distillation process.

    \item[(9)] \textbf{Transfer Learning with Pruning and Adapter (TL+P+A):}
    In this approach, the pruned and transferred, TL+FT+Q, models, undergo architectural modifications with the addition of an adapter layer. This leverages the concept of depthwise separable convolution to minimize its size and is exclusively trained on the target data, while all other layers remain frozen.

    \item[(10)] \textbf{Transfer Learning with Knowledge Distillation and Adapter (TL+KD+A):} This approach builds on the distilled models used for transfer learning, as outlined in TL+P+FT, by incorporating adapter layers into their architecture. While these adapters are identical to those described in method (8), their placement differs: instead of being positioned in the middle of the layers, as in the pruned approach, they are added after the last convolutional layer (specific to each architecture). The adapter is then trained independently.
\end{enumerate}

These approaches are different in terms of performance, training time, and inference time. Taking into consideration that synthetic images in a dataset can be generated from  multiple types of generative models (i.e., GANs, VAEs, diffusion models), serve different purposes (i.e., entertainment, realism enhancement, data augmentation), and have different levels of manipulations (i.e., splicing, object removal, face swapping), we experimentally compared them with three datasets as it will be described in Section \ref{sec:ExperimentalEvaluation}.





\section{Experimental Evaluation}
\label{sec:ExperimentalEvaluation}

\subsection{Experimental Setup}

We used three datasets for our experiments. The first  comprises DeepFake images from the SynthBuster test set \cite{bammey2023synthbuster} \footnote{https://zenodo.org/records/10066460} and authentic images from the RAISE dataset \footnote{http://loki.disi.unitn.it/RAISE/download.html}. 
The SynthBuster contains 9,000 AI-generated images (1,000 each) from nine DeepFake models: DALL·E 2, DALL·E 3, Adobe Firefly, Midjourney v5, and five versions of Stable Diffusion. RAISE contains 8,153 authentic images.
Images are split into 60\% for training and 40\% for testing, with both sets containing a mix of authentic images and those generated by all nine DeepFake models.

The second dataset is ForenSynths  \cite{wang2020cnn} \footnote{https://github.com/PeterWang512/CNNDetection}, which includes authentic images and DeepFake images generated from 13 different CNN-based GAN models.
The training set contains 720,119 images, the validation set 8,000, and the test set 90,310. For our experiments, we focused exclusively on human face images, as deepfakes are often used to impersonate prominent individuals, leading to potential harm, confusion, and security threats \cite{verdoliva2020media}, \cite{westerlund2019emergence}, \cite{misirlis2023deepfake}. It is important to emphasize that the DeepFake images in the training set are exclusively generated by the ProGAN model, while the test set includes images not only from ProGAN but also from other models like StarGAN and WhichFaceIsReal, which are absent from the training set.

To evaluate the transfer learning approach, we utilized a simple ''dogs vs. cats'' dataset
\footnote{https://www.kaggle.com/datasets/salader/dogs-vs-cats}
consisting of 25,000 images to train the original models. These were later employed to accelerate the training process using techniques such as Fine-Tuning, KD, and adapter layers.
All images from the datasets were resized to $224 \times 224$ pixels. Experiments were conducted in Python 3, utilizing libraries such as PyTorch, PIL, and scikit-learn. They were executed on Kaggle notebooks, using two NVIDIA T4 GPUs, each with 16 GB of memory.

The experiments involving the aforementioned datasets utilized a primary VGG-based model consisting of approximately 4.5 million parameters. The code for the primary BL model, the compressed and transferred models  are available in \cite{karathanasis}.



\subsection{Outcomes}

\begin{figure}[ht!]
    \centering
    \includegraphics[width=1\linewidth]{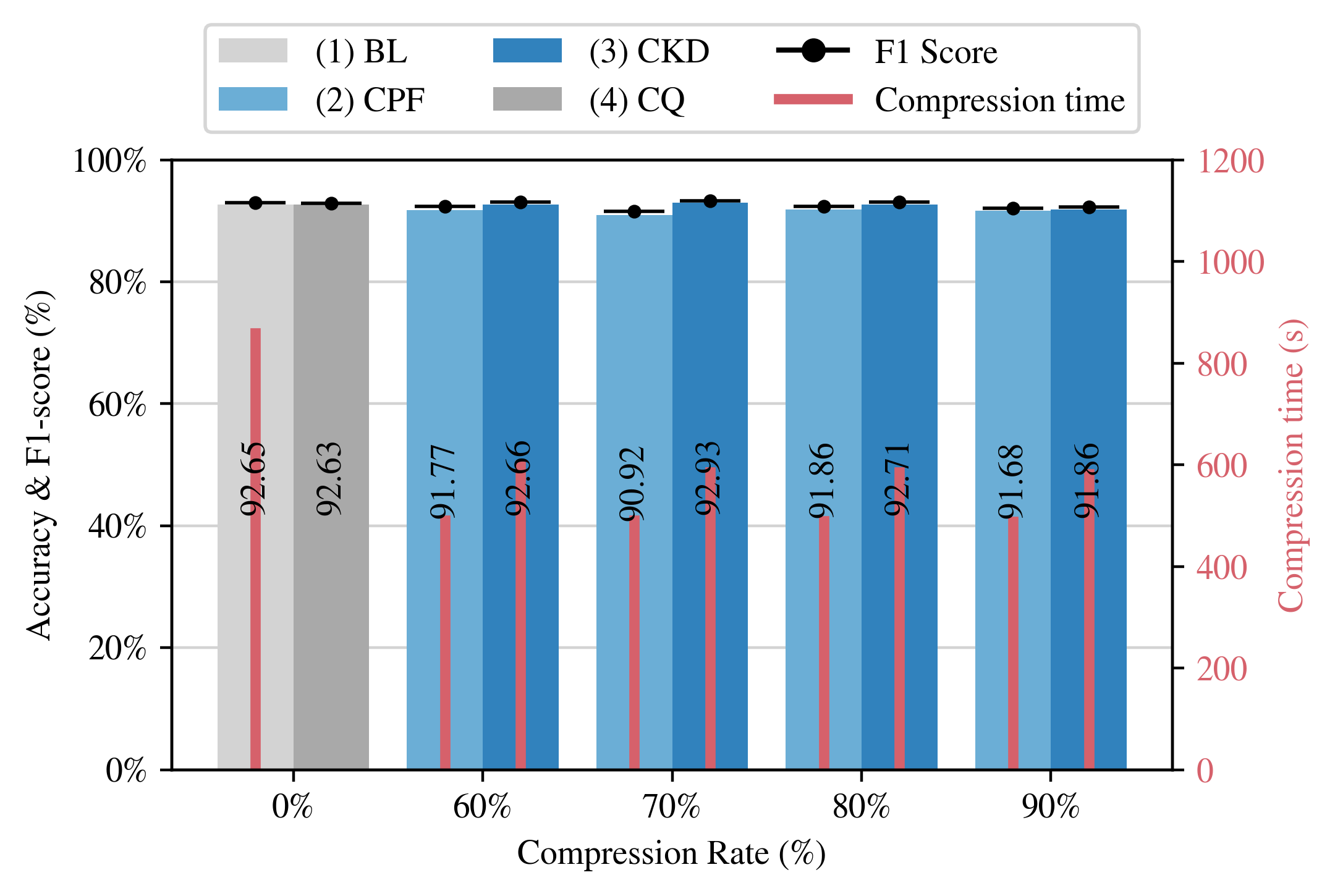} 
    \caption{Evaluation of Compression Techniques Using a Dataset Combining Deepfake Images from Synthbuster and Authentic Images from RAISE}
    \captionsetup{justification=centering}
    \label{fig:Synthbuster}
\end{figure}

\begin{figure*}[ht!]
    \centering
    \includegraphics[width=1\linewidth]{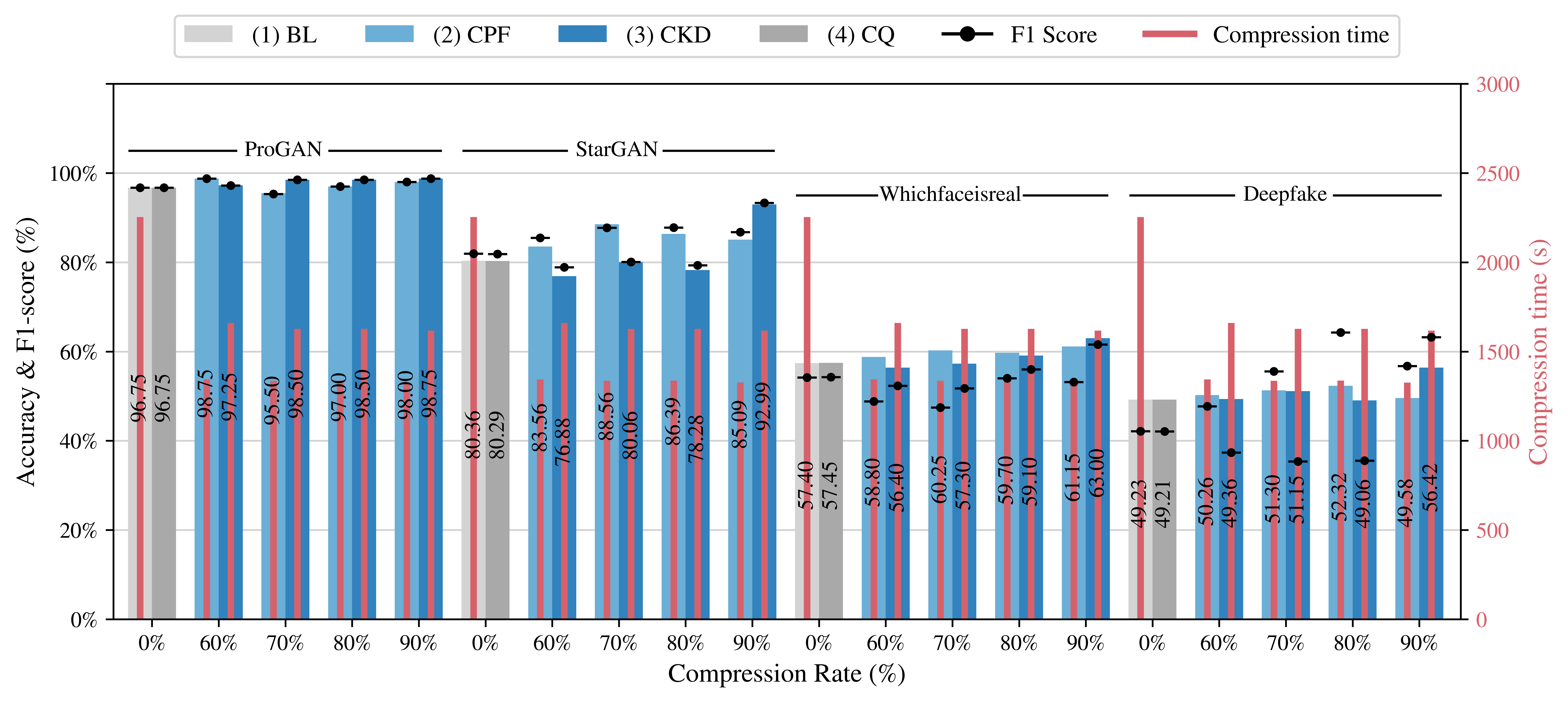} 
    \caption{Evaluation of Compression Techniques Using ForenSynths dataset }
    \captionsetup{justification=centering}
    \label{fig:ForenSynths_Comp}
\end{figure*}

\begin{figure*}[ht!]
    \centering
    \includegraphics[width=1\linewidth]{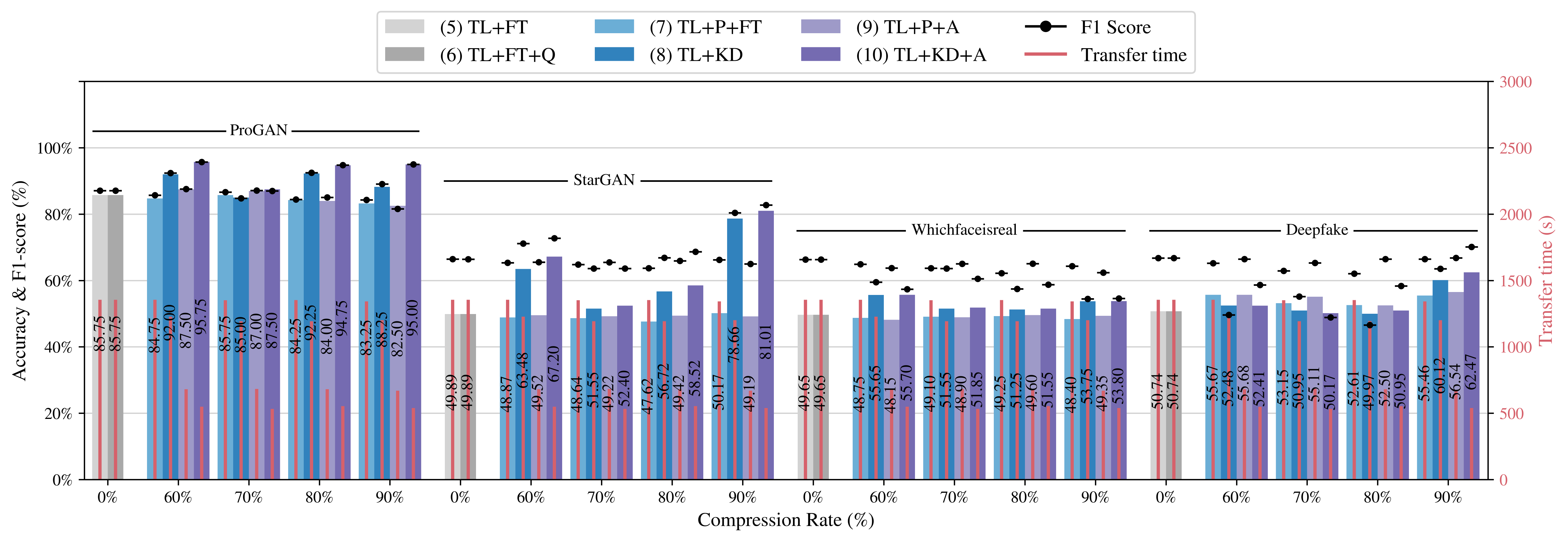} 
    \caption{Evaluation of Transfer Learning Techniques Using ForenSynths dataset }
    \captionsetup{justification=centering}
    \label{fig:ForenSynths_Transfer}
\end{figure*}

Figures \ref{fig:Synthbuster} and \ref{fig:ForenSynths_Comp} present the results of the experiments using compression approaches, while Figure \ref{fig:ForenSynths_Transfer} illustrates the experiments with transfer learning approaches. The acronyms in the legend above the box plots correspond to the methods outlined in Section \ref{sec:ProposedMethodology}. Evaluations are presented across compression rates ranging from 60\% to 90\% relative to the original model size in terms of parameters. We present the Accuracy, F1-Score, compression time and transfer learning time. We do not provide compression time for the quantization,  as converting from float to int is almost negligible.
The red lines in the Baseline methods indicate the training time of the original models.
Figure \ref{fig:Synthbuster} presents the outcomes on the Synthbuster dataset. The findings reveal that the pruned models achieved performance comparable to the original larger model. However, KD demonstrated slightly superior results, with some student models even outperforming the teacher model (baseline approach). 

Figures \ref{fig:ForenSynths_Comp} shows the results obtained on the ForenSynths dataset. Since the test images were generated by different synthetic image generators, we evaluated the images from each generator separately. When testing with images generated by ProGAN, which also produced the training dataset, all methods achieved very high accuracy, even at high compression levels. However, performance declines with images from StarGAN, and the degradation becomes more pronounced with WhichFaceIsReal and DeepFake generators. This occurs because StarGAN, like ProGAN, is also a GAN-based model with several similarities; however, the synthetic generation processes used in WhichFaceIsReal and DeepFake differ significantly. Nonetheless, it is worth noting that the compression methods, including pruning, KD, and quantization, deliver accuracy comparable to the uncompressed baseline models across all cases.


Similarly, Figure \ref{fig:ForenSynths_Transfer} present the evaluation results of transfer learning using the ForenSynths dataset. Testing on the set generated by ProGAN shows strong performance, whereas the other test sets exhibit significantly lower performance, highlighting a domain generalization challenge. When comparing methods, KD generally outperforms pruning and incorporating an adapter typically enhances performance in most cases.

\subsection{Discussion}


In Compression approaches we observed that even if the compression level is very high, reaching up to 90\% compared to the initial size, in all cases, the accuracy of compressed models remain at the same level with the baselines. Furthermore, we observe that when the testing and training datasets come from the same deepfake generator, in almost all the aces, KD surpasses the pruning approach, while when we use a different deepfake generator the pruning approach leads to better results in all cases except from the very high compression level of 90\%.

Quantization effectively maintained performance but did not achieve the same level of compression as KD and pruning. Additionally, quantized DeepFake models cannot benefit from accelerators like GPUs, limiting their practical advantage. In some cases, depending on the hardware, quantization might not have any benefit at all. However, for DeepFake detection on Android devices for example, where inference is typically done on optimized mobile CPUs, quantization remains valuable.

It is also worth noting that we conducted experiments with low-rank factorization, supported as a method by \cite{masana2017domain}, which reduced model size to some extent but often caused significant performance degradation, limiting its utility. That is also the reason that the method's results are not included in this paper. Its effectiveness is context-dependent, and the compression achieved usually requires combining it with other methods to meet the needs of DeepFake edge computing.

Regarding transfer learning, we found that the adapter's position significantly affects performance. In deeper models, placing the adapter near the first or middle layers maximized effectiveness. In smaller models, such as student models in KD, placing the adapter at the end of the convolutional layers yielded the best results. Notably, adapters only improved feature extraction; those using linear layers showed no performance gains in some preliminary experiments we conducted.

\section{Conclusion \& Future Work}
\label{sec:Conclusions}
In this paper we observe that even with a high-level compression of nearly 90\% of the original model, performance remains consistent when the testing dataset originates from the same DeepFake generator as the training/compression dataset. However, there is a significant drop in performance when the evaluation dataset is generated by a different DeepFake generator. This highlights a domain generalization challenge combined with compression in DeepFake detection, which we aim to address in our future research.

\section*{Acknowledgment}
This work was funded by the European Union’s Horizon Europe research and innovation program under grant agreement No. 101120237 (ELIAS).



\balance
\bibliographystyle{IEEEtran}
\bibliography{references}

\end{document}